# A Grey-box Launch-profile Aware Model for C+L Band Raman Amplification

Yihao Zhang[(1)], Xiaomin Liu[(1)], Yichen Liu[(1)], Lilin Yi[(1)], Weisheng Hu[(1)], Qunbi Zhuge*[(1)]

[(1)] State Key Laboratory of Advanced Optical Communication Systems and Networks, Department of Electronic Engineering, Shanghai Jiao Tong University, Shanghai 200240, China, qunbi.zhuge@sjtu.edu.cn

**Abstract** *Based on the physical features of Raman amplification, we propose a three-step modelling scheme based on neural networks (NN) and linear regression. Higher accuracy, less data requirements and lower computational complexity are demonstrated through simulations compared with the pure NN-based method.* ©2022 The Author(s)

**Introduction**

Raman amplification (RA) attracts increasing attention for its ability of providing wide gain profiles with low noise in multi-band transmission systems [1]. The configuration of Raman pumps determines the shape of the gain profile, which in turn affects the quality of transmission. In multi-band systems, the stimulated Raman scattering (SRS) between signals and between signals and pumps can make the effect of RA on the signals much more complicated. Therefore, the optimization of the configuration of Raman pumps is required, for which an accurate model of RA is needed. A sufficiently accurate model is the theoretical model based on a set of ordinary differential equations (ODEs) describing the SRS [2]. However, solving the ODEs is time-consuming [3], which cannot satisfy the low-latency requirement of applications such as efficient resource allocation and automatic optimization for future network management. To predict the gain profile of RA faster, machine learning (ML) is one of the popular methods.

The neural network (NN) is an attractive ML tool for the modelling of RA due to its strong learning ability and parallel computing structure. In [3-5], NNs are utilized to build mapping from a set of Raman pump power to the gain profile. However, the existing studies only consider modelling with fixed launch power and flat power profiles. Such modelling schemes are not applicable to networks with complex and dynamic link conditions. In practical scenarios, the launch power profile needs to be considered when modelling the RA. However, if the launch power profile is directly added to the input layer of the NN, the input space will become enormous, resulting in a dramatic increase in the number of NN parameters. This can lead to problems such as high computational complexity of the NN, overfitting on the training data-set, and requirements of generating huge training data-sets. Since each Raman amplifier needs modelling individually in real networks, acquiring a large data-set for each amplifier would be time-consuming. Thus, how to achieve higher modelling accuracy with less data is an important issue in practice.

To solve the abovementioned problems, in this paper, a grey-box modelling method for C+L band RA considering the launch power profile is proposed. A three-step modelling scheme based on NN and linear regression (LR) is proposed, where physical features of RA are leveraged. Simulation results show that the proposed scheme can lower the computational complexity by reducing 46.7% NN parameters, and meanwhile reduce the prediction error from 0.56 dB to 0.15 dB with a small data-set compared with the conventional method. Additionally, the proposed scheme reduces the maximum error by ~0.7 dB, showing the ability to alleviate the overfitting problem and improve the reliability.

**Principle**
*Prediction Process*
First, the process of predicting RA gain profiles for signals with arbitrary launch power profiles using the proposed grey-box modelling scheme is described in Fig. 1(a), which contains three steps:

i) The launch power profile is neglected and the powers of Raman pumps are input to a base model which is an NN to build the following mapping:

$$f_{\text{NN1}}: \mathbf{P}_{\text{pump}} \rightarrow \mathbf{G}_{\text{base},@P_{\text{ref}}}, \qquad (1)$$

where $\mathbf{P}_{\text{pump}}$ denotes the vector of the powers for all pumps. The output of the NN is $\mathbf{G}_{\text{base},@P_{\text{ref}}}$, which denotes the Raman net gain profile predicted by this base model. The subscript $@P_{\text{ref}}$ means this NN is trained in the case where the total launch power is equal to $P_{\text{ref}}$. The net gain is a commonly used metric, defined as the difference between the output optical power of the fiber and the launch optical power [2].

ii) The power difference between the total launch power $\Sigma P_{\text{sig}}$ and $P_{\text{ref}}$ is calculated. Then, together with $\mathbf{P}_{\text{pump}}$, it is input to a second NN:

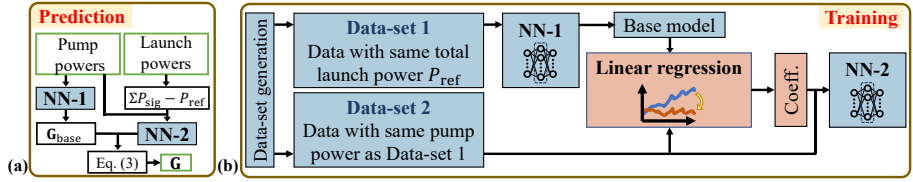

**Fig. 1:** The proposed grey-box training and prediction scheme for Raman amplification.

$$f_{\text{NN2}}: (\mathbf{P}_{\text{pump}}, \Sigma P_{\text{sig}} - P_{\text{ref}}) \to (b_1, b_2, b_3), \quad (2)$$

where the outputs $b_1$, $b_2$, and $b_3$ are three linear regression coefficients, which are used for the next step for model transfer.

iii) The $\mathbf{G}_{\text{base},@P_{\text{ref}}}$ is transferred to a final predicted gain profile $\mathbf{G}$ at a given launch power profile through:

$$\mathbf{G} = b_1 + b_2 \cdot \mathbf{n}_{\text{ch}} + b_3 \cdot \mathbf{G}_{\text{base},@P_{\text{ref}}}, \quad (3)$$

where $\mathbf{n}_{\text{ch}}$ is a vector of $[1:1:N_{\text{ch}}]$ and $N_{\text{ch}}$ is the total number of channels. Eq. (3) leverages the similarity of the shapes of the Raman gain profiles with same Raman pumps but with different total signal launch power, which will be detailed later.

*Training Process*

The training process of the proposed grey-box modelling scheme is shown in Fig. 1(b).

**i) Training NN-1:** To build the base model (the first NN), a data-set is generated. To address the problem that considering the launch power profile may lead to too high dimension of the input space, we fix the total launch power to be $P_{\text{ref}}$ when building the base model. Thus, data are collected when total launch power equals to $P_{\text{ref}}$. A ripple up to 3 dB of the launch power profile is considered. The ripple is defined as the difference between the maximum and minimum values of launch power per channel. This is because the launch power profile usually fluctuates randomly around a certain power level in practice [6]. Then the first NN is trained using the data-set. Ignoring the launch power profile here can reduce the dimension of the input layer, simplify the NN structure of the base model, and reduce the training burden.

**ii) Training NN-2:** To train the second NN, the labels $b_1, b_2$, and $b_3$ must be obtained in advance. Thus, sufficient pairs of $(\mathbf{G}, \mathbf{G}_{\text{base},@P_{\text{ref}}})$ in Eq. (3) is generated, and the coefficients $b_i$ are fitted through linear regression. Because $\mathbf{G}_{\text{base},@P_{\text{ref}}}$ has been generated and the two profiles $\mathbf{G}$ and $\mathbf{G}_{\text{base},@P_{\text{ref}}}$ correspond to the same $\mathbf{P}_{\text{pump}}$, the second data-set is designed to contain data with the same Raman pump power as those in the first data-set, but the total launch power is different. Due to the low computational complexity of the linear regression, this step can be performed at minimal cost. Then the second NN is trained using the data in the second data-set as features and the fitted $b_i$ as the labels.

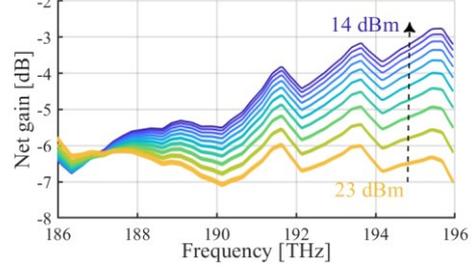

**Fig. 2:** The Raman net gain profiles in the cases of 10 different total launch power levels. 500 random launch power profiles with 3-dB ripple are generated for each power level. (Each cluster contains 500 curves.)

*Physical Interpretation*

The rationality of the proposed scheme is illustrated below. In Fig. 2, the Raman net gain profiles provided by the same Raman pumps for 10 total launch power levels ranging from 14 dBm (-9 dBm per channel) to 23 dBm (0 dBm per channel) with a step of 1 dB are shown. For each power level, 500 profiles with 3-dB ripple are randomly generated. Although the shape of the launch power profile may vary considerably within 3 dB, the fluctuation of gain profiles is very small (within 0.008 dB). This indicates that for a fixed total launch power with limited ripple, the shape of the power profile has almost no effect on modeling accuracy. Fig. 2 also indicates that variations of the total launch power result in relatively significant changes in the gain profile, but the gain's shape is always similar. As a result, Eq. (3) is established for model transformation, where the three terms to the right of the equation represents the offset, tilting, and scaling between two profiles with different total launch powers, respectively.

**Simulation Results and Discussions**

The data-set generation is based on GNPy [7], an open-source library for building route planning and optimization tools in optical networks. The modelling for a Raman amplifier on an 80 km long G.652D standard single mode fiber (SSMF) is considered. The number of Raman pumps is set to be 5 to provide a gain across the entire C+L band, spanning 10 THz from 186 THz to 196 THz. The wavelengths of these Raman pumps are 1426 nm, 1440 nm, 1454 nm, 1472 nm, and 1496 nm. 200 channels of signals are launched with a baudrate of 35 GBaud and a channel spacing of 50 GHz. For the first data-set, the total launch power $P_{\text{ref}}$ is set to 14 dBm and the launch power ripple is set to 3 dB. The power of each Raman pump subjects to the uniform distribution of $\sim \mathcal{U}(0, 0.2)$ W. For the second data-set, the total launch power is randomly chosen from 15 dBm to 23 dBm with a step of 1 dB, and the power ripple is also 3 dB. The first NN has a hidden-layer with 200 nodes, and the second NN has two hidden layers with 50 and 30 nodes, respectively.

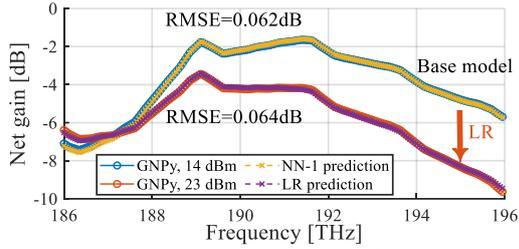

**Fig. 3:** An example of the net gain profiles calculated by GNPy and the predicted profiles. The gain profile when the total launch power is 14 dBm is predicted by the first NN, then it is transferred to the 23-dBm case through the LR. (Due to the limited pump power, the net gain is negative. The residual attenuation can be compensated for by EDFAs in practice.)

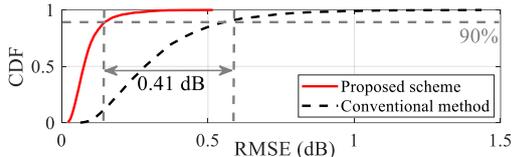

**Fig. 4:** The CDF the proposed scheme and the conventional method. The data-set size is 500 for both cases.

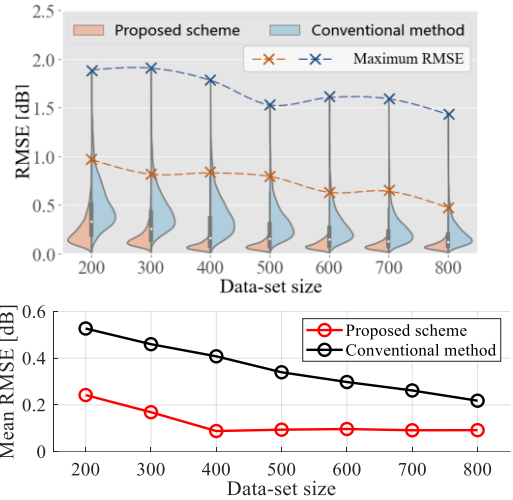

**Fig. 5: (a)** The RMSE distribution and **(b)** the mean RMSE over the validation set, for both the proposed scheme and the conventional method with different data-set sizes.

For comparison, simulations on the conventional NN-modelling method are conducted, where the launch power profile is added to the input-layer of the NN. The number of nodes in the hidden-layer is also 200. For both the proposed scheme and the conventional method, a maximum of 800 data are generated for training and testing (80% for training and 20% for testing). Another independent data-set is generated for the validation of both methods.

First, the error of the base model and the model transferring through linear regression is presented, which is defined by the root mean square error (RMSE) between the predicted and the true profiles. An example when the pump powers are 0.01, 0.16, 0.12, 0.19, and 0.13 W is shown in Fig. 3. The prediction RMSE of the first NN for establishing the base model is 0.062 dB, and the prediction RMSE of linear regression for model transfer is 0.064 dB. The results show that both the base model and the transferred model can achieve high accuracy. Next, in Fig. 4, the cumulative distribution function (CDF) for RMSE of the proposed modelling scheme is shown, and the CDF for RMSE of the conventional method is plotted for comparison. Results show that the proposed scheme can reduce the RMSE from 0.56 dB to 0.15 dB when the CDF reaches 90%.

To evaluate the performance when training with different data-set sizes, in Fig. 5(a) the RMSE distribution over the validation set of the proposed scheme and the conventional method with different amount of data is plotted. Results show that the proposed scheme always has a more concentrated RMSE distribution, and can reduce the maximum RMSE over the validation set by at least ~0.7 dB with a data-set size of 500. This indicate that the proposed scheme is more reliable and can overcome the problem of overfitting on the training set to some extent. Moreover, as shown in Fig. 5(b), the mean RMSE over the validation set of the proposed scheme is smaller, indicating that the proposed scheme is more accurate. This also suggests that the proposed scheme makes it possible to model RA with small data-sets, being more convenient for practical utilization.

Moreover, the conventional method has 81400 NN parameters, while the proposed scheme has only 43373 parameters (46.7% reduction), which reduces the computational complexity for training and prediction. In conclusion, the proposed scheme can improve the accuracy and reliability, reduce the amount of training data, and lower the computational complexity of the RA modelling.

**Conclusion**

A grey-box modelling scheme for RA considering the launch power profile is proposed. We first establish a base model considering certain launch power, and then transfer the base model to other cases with arbitrary launch power profiles. The proposed scheme achieves higher accuracy and reliability with lower computational complexity and smaller training data-set compared to the conventional method.

**Acknowledgements**

This work was supported by the National Key R&D Program of China (2018YFB1800902) and NSFC (62175145).